# Physics-Informed Neural Network for Predicting Out-of-Training-Range TCAD Solution with Minimized Domain Expertise

Albert Lu
Electrical Engineering
San Jose State University
San Jose, USA

Yu Foon Chau
Mathematics Department
University of California, Irvine
Irvine, USA

Hiu Yung Wong*
Electrical Engineering
San Jose State University
San Jose, USA

*Abstract*— Machine learning (ML) is promising in assisting technology computer-aided design (TCAD) simulations to alleviate difficulty in convergence and prolonged simulation time. While ML is widely used in TCAD, they either require access to the internal solver, require extensive domain expertise, are only trained by terminal quantities such as currents and voltages, and/or lack out-of-training-range prediction capability. In this paper, using Si nanowire as an example, we demonstrate that it is possible to use a physics-informed neural network (PINN) to predict out-of-training-range TCAD solutions without accessing the internal solver and with minimal domain expertise. The machine not only can predict a 2.5 times larger range than the training but also can predict the inversion region by only being trained with subthreshold region data. The physics-informed module is also trained with data without the need for human-coded equations making this easier to be extended to more sophisticated systems.

## I. INTRODUCTION

Machine learning (ML) is promising in assisting technology computer-aided design (TCAD) simulations to alleviate difficulty in convergence and prolonged simulation time. Roughly, ML is used in TCAD in three approaches. In the first approach, ML is used to improve solver performance by providing an initial guess [1]-[3]. This requires access to the internal solver of a TCAD tool and significant domain expertise by coding physical equations (including differential operators) in the loss functions. The second approach is to use TCAD to generate terminal data such as currents and voltages and then use ML to perform reverse engineering (such as defect identification [4]-[7]) or simulation emulation [8]-[10]. This does not require access to the internal solver and has been proven to be able to predict out-of-training range data [10]. If an appropriate machine learning method is used, it can also minimize the requirement for domain expertise [8]-[10]. The third approach is to solve the TCAD problem by training a machine with the spatial distribution of physical quantities such as electron density and potential [11][12]. However, most of the works demonstrated so far only train a machine using the physical quantities without the knowledge of the physics of the device (as the solver is not accessible) and only use it to predict the physical quantities within the training range.

Physics-informed neural network (PINN), in which the physics of the problem is incorporated into the NN, is expected to improve predictability [13]. In this paper, we study a new approach in which a machine is trained by the spatial distribution of physical quantities with PINN but with minimal domain expertise. A commercial TCAD tool in which the solver is not accessible is used. The purpose is to study the possibility of out-of-training-range prediction using such a machine. Si nanowire is used as an example.

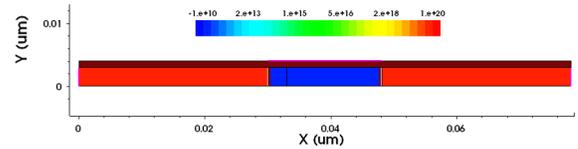

Fig. 1: TCAD generated structure used in this study (doping concentration is shown).

## II. DATA GENERATION

An n-type silicon nanowire is studied using TCAD Sentaurus. Half of the domain was simulated due to its symmetry. Fig. 1 shows the structure with a radius of 4nm, gate length of 18nm, and oxide thickness of 1nm. The source and drain region are doped with $10^{20}$ cm$^{-3}$ of arsenic and the body is doped with $10^{10}$ cm$^{-3}$ of boron (~undoped). Tensor mesh is used with a size of 129 × 17 (2193 mesh points). The electrostatic potential ($\phi$), electron density ($n$), and net space charge after device simulation at each point are extracted using a script written for TDX in TCAD Sentaurus. The extracted quantities are verified with a Python-coded Poisson equation.

The structure is then simulated in SDevice with a setup that includes Fermi-Dirac statistics and high field saturation. It is run as a quasistationary simulation and only the Poisson equation is solved with the drain and source voltages at 0 V. The gate voltage, $V_G$, is ramped from 0 to 0.75 V, and snapshots of the electrostatic potential and electron density are taken for every 7.5 mV for a total of 101 snapshots. Note that each 'snapshot' (profile) records $\phi$ and $n$ distributions in the 2193-point mesh.

---

*Corresponding author: hiuyung.wong@sjsu.edu

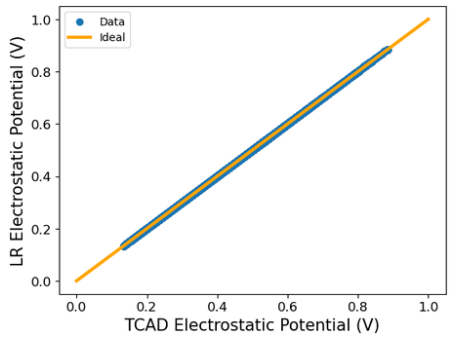

Fig. 2. Comparison between the electrostatic potential predicted using the physics-informing LR module and the electrostatic potential from TCAD. The blue points consist of all mesh points at all $V_G$ from 0V to 0.75V. Note that the LR model is only trained on the first 40 $V_G$ from 0V to 0.3V. The prediction matches the ideal case.

### III. PHYSICS-INFORMING MODULE

Firstly, a physics-informing module (not PINN yet) formed by a machine is created. This machine needs to predict the $\phi$ profile for a given $n$ profile. For this problem, since $n$ dominates the space charge, this is equivalent to making a Poisson solver for a given net space charge ($\rho$) distribution,

$$\nabla \cdot (\varepsilon \nabla \phi) = -\rho \quad (1)$$

where $\varepsilon$ is the permittivity and $\rho \sim n$.

It is found that training a machine using linear regression (LR) and the first 40 snapshots ($V_G$=0V to 0.3V) is sufficient for this task. The input is the $n$ profile and the output is the $\phi$ profile. To make it easier to train, the electron density is added by $10^{10}$ and normalized by $10^{19}$. $10^{10}$ allows the mesh points in the oxide region to be nonzero and thus be able to be taken the logarithm of. $10^{10}$ is chosen since it is still found to be insignificant considering the scale of the other mesh points are orders of magnitude larger. Fig. 2 plots $\phi$ predicted by the machine against the $\phi$ calculated by TCAD for every mesh point from $V_G$=0V to 0.75V for a given $n$ profile from TCAD and they agree with each other. Note again the machine has not seen the profiles for $V_G > 0.3V$. This demonstrates the ability of the machine to learn the physics because the electrostatic potential in the channel varies significantly and non-linearly while those in the source and drain stay almost constant. Moreover, the training data are in the subthreshold regime which has a very different $V_G$ dependency than the inversion regime. LR is sufficient because when the charge profile is given, Eq. (1) is just a linear matrix multiplication after discretization. Therefore, the machine is expected to model the inverse of the matrix at the left-hand side of the problem. Systems of differential equations are generally discretized and linearized. Therefore, this methodology is expected to be applicable to more sophisticated problems.

### IV. PINN ARCHITECTURE

A PINN is then built. It consists of a convolutional neural network (CNN) and the aforementioned physics-informing module. Fig. 3 shows the architecture. The input is the desired $V_G$ and the output is both the electron density profile and electrostatic potential profile. For any given $V_G$, the $n$ profile is predicted by a CNN. The output layer of the CNN is chosen to be the exponential linear unit (ELU) to allow for easily taking the logarithm of the value later. As a post-processing step, the $n$ profile is then added by 1 and $10^{-9}$. 1 is added because ELU has a minimum prediction of -1 and thus this takes the minimum value to 0. However, the logarithm of 0 is still undefined so $10^{-9}$ is added to keep it slightly above 0. $10^{-9}$ is also used because the LR module uses $(n + 10^{10})/10^{19}$. Thus, $10^{10}/10^{19}$ is $10^{-9}$ when $n$ is small. This ensures that the equations are balanced properly so that the proper values are learned. It will then be passed to the physics-informing LR module which will calculate the corresponding $\phi$ profile, from which the gate voltage, $V_G'$, can be extracted. If the solution is correct, $V_G' = V_G$. This comparison forms the first loss function and enforces the gate boundary condition.

The predicted $\phi$ profile is then further used to predict the $n$ profile using the Fermi-Dirac (FD) module in Fig. 3, which is also a physics-informing module but much more straightforward. It calculates the $n$ profile based on the $\phi$ profile through the approximation of the Fermi Integral of order ½ for Fermi-Dirac statistics [14] and is compared to the n profile predicted by the CNN. Note that the FD module also adds $10^{10}$ and normalizes by $10^{19}$ in order to balance the additions added in the other modules. This forms the second constraint and is used to ensure that the predicted electron density obtained through the approximate Fermi Integral matches the predicted electron density from the CNN. This essentially enforced the Fermi-Dirac statistics. Fig. 4 shows that the approximate Fermi Integral calculation is correct by plotting the approximation against the actual value obtained from TCAD. However, in contrast to the first loss, the logarithm of both the predicted electron density and the one obtained through the approximation are taken and then MSE is applied. This is found critical to ensure that the PINN learns $n$ appropriately across several orders of magnitude ($10^{10}$ to $10^{21}$). Without using the logarithm, it was found that the losses would be prioritized at high $n \sim 10^{19}$ or greater. This is due to the normalization of $n$ by $10^{19}$ which would make the values less than $10^{19}$ essentially zero. This decreases the contribution of the lower $n$ values to the overall prediction.

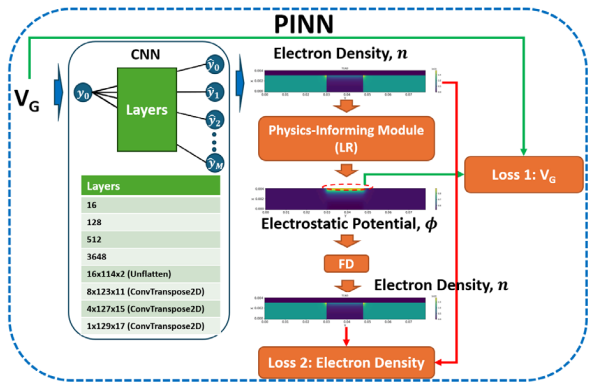

Fig. 3. The machine used in this study consists of a CNN and 2 loss function. The CNN layers are detailed in the table. $V_G$ is input to the PINN and the electron density is the output. The physics-informing module (LR machine) is trained only using the first 40 $V_G$ (0V to 0.3V) snapshots of the electrostatic potential and electron density. The predicted electron density is passed through the LR machine to obtain the electrostatic potential. The gate contact electrostatic potential is expected to match the input $V_G$ through a loss function. The predicted electrostatic potential is also passed through the approximation of the Fermi Integral of order ½ (FD) to obtain the electron density. Another loss function is used to ensure it matches the predicted electron density from the CNN.

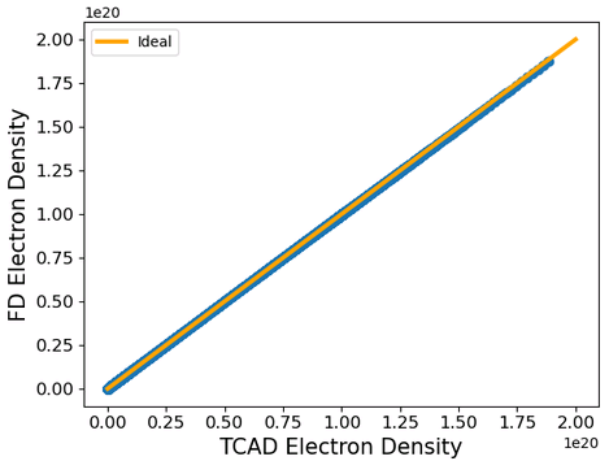

Fig. 4. Comparison between the electron density predicted using the approximate Fermi-Dirac (FD) Distribution (Fermi Integral of order ½) and the electron density from TCAD. The blue points consist of all mesh points at all $V_G$ from 0V to 0.75V.

For each $V_G$ of interest, the two losses will be minimized using the Adam optimizer. To speed up the process, the learning rate is initially $10^{-3}$ but as it plateaus it is changed adaptively until a minimum of $10^{-5}$ is reached. This process can be treated as a replacement for TCAD solvers. This can also be regarded as *self-supervised learning*. Note again that, in the whole process, only the LR module has been trained with the first 40 $V_G$ snapshots. Moreover, unlike regular TCAD simulation, one can calculate $V_G = 0.75V$ directly without ramping $V_G$.

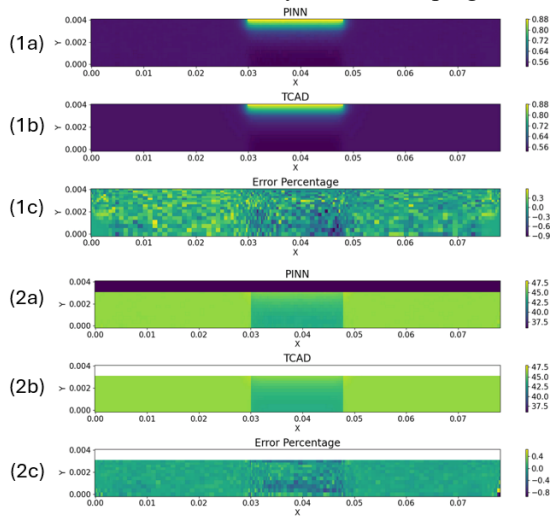

Fig. 5. Predicted electrostatic potential profile using the PINN (1a) and the electrostatic potential from TCAD (1b) when $V_G$=0.75V. The max error percentage between them is less than 0.3% (1c). Predicted electron density profile in log scale using the PINN (2a) and the electron density from TCAD in log scale (2b) when $V_G$=0.75V. The max error percentage in log scale between them is less than 0.6% (2c). Note that the PINN has only seen the first 40 $V_G$ snapshots ($V_G$=0V to 0.3V). Note that (2a) had the oxide region added by $10^{10}$ but it is essentially zero.

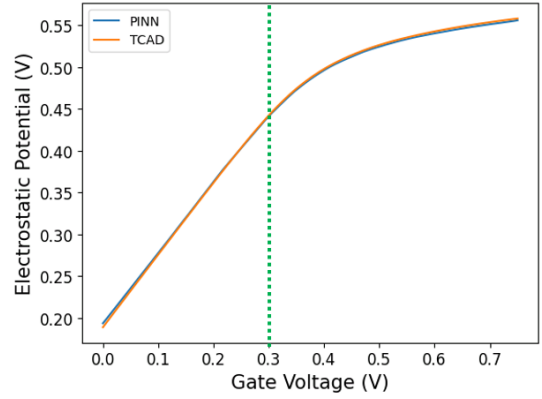

Fig. 6. Comparison of one of the points in the mesh at x=0.0405um and y = 0.002um where the electrostatic potential from the PINN and TCAD as the $V_G$ is ramped from 0V to 0.75V. The green dotted line indicates that $V_G$=0.3V is where the physics-informing LR module was trained up to. This mesh point is under the gate region.

## V. RESULTS

Fig. 5 shows the predicted $n$ profile and $\phi$ profile compared to TCAD simulation results for $V_G$. The maximum error is less than 0.3% for the $\phi$ profile and less than 0.6% for the $n$ profile in the logarithmic scale. Fig. 6 shows the prediction of a point in the middle of the structure under the gate region across all $V_G$. This point was chosen because it experiences the transition from the depletion to the inversion regime. The result matches the TCAD simulation well. This demonstrates the ability of the PINN to learn the physics of this device and predict out-of-training-range results. It is not based on simple extrapolation. Fig. 7 further shows the predicted vs. actual $\phi$ and $n$ of every mesh point across all $V_G$. The $n$ comparison is in log scale and shows how the PINN is able to learn across several orders of magnitude successfully.

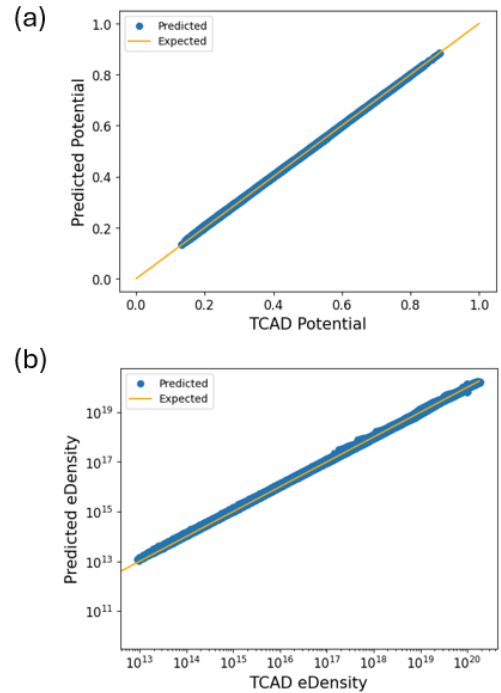

Fig. 7. (a) shows the comparison of the PINN-predicted $\phi$ profile and the TCAD $\phi$ profile in linear scale. (b) shows the comparison of the PINN-predicted $n$ profile and the TCAD $n$ profile in log scale.

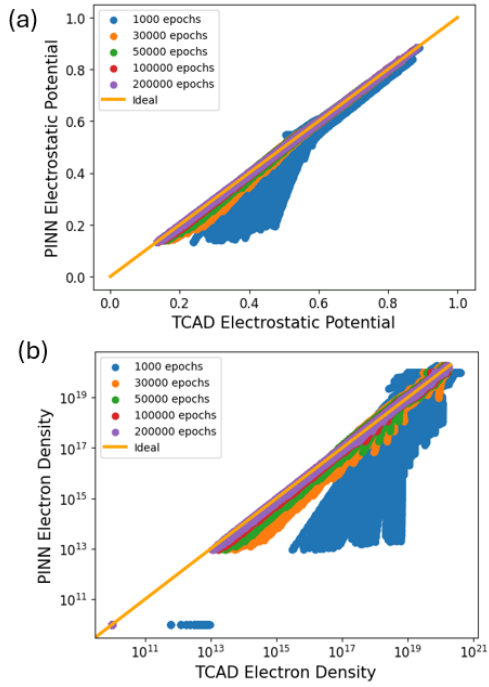

Fig. 8. (a) shows the comparison of the PINN-predicted $\phi$ profile and the TCAD $\phi$ profile with respect to the number of epochs. (b) shows the comparison of the PINN-predicted $n$ profile and the TCAD $n$ profile in log scale with respect to the number of epochs.

The number of epochs in the previous figures is 200,000. However, further observations are made between the number of epochs and the performance. Fig. 8 shows that a lower number of epochs may still provide sufficiently good results. It can be seen that 100,000 epochs is close to 200,000 epochs. Furthermore, as low as 30,000 epochs may be enough if a lower accuracy is acceptable.

## VI. CONCLUSIONS

A novel framework based on a PINN is demonstrated to have effectively learned the underlying physics by using specific physics-based loss functions. It can predict out-of-training-range simulation results as it has learned the underlying physics. A machine trained by the TCAD data from $V_G$ = 0V to 0.3V (below the subthreshold region) is shown to be able to predict the $\phi$ and $n$ profiles for any given $V_G$ (up to 0.75V).

ACKNOWLEDGMENT

Part of the work was supported by the National Science Foundation under Grant No. 2046220.